\documentclass[pdflatex,sn-mathphys-num]{sn-jnl}

\usepackage{graphicx}%
\usepackage{multirow}%
\usepackage{amsmath,amssymb,amsfonts}%
\usepackage{amsthm}%
\usepackage{mathrsfs}%
\usepackage[title]{appendix}%
\usepackage{xcolor}%
\usepackage{textcomp}%
\usepackage{manyfoot}%
\usepackage{booktabs}%
\usepackage{algorithm}%
\usepackage{algorithmicx}%
\usepackage{algpseudocode}%
\usepackage{listings}%

\theoremstyle{thmstyleone}%
\theoremstyle{thmstyletwo}%

\theoremstyle{thmstylethree}%

\begin{document}

\title[Engineering-Oriented HD Map generation Workflow]{Geo-Data-Driven HD Map Generation Workflow with Integrated Reference-Free Constraint-Based Verification}

\author*[1]{\fnm{Ruidi} \sur{He}}\email{ruidi.he@tu-clausthal.de}
\author[1]{\fnm{Vaibhav} \sur{Tiwari}}\email{vaibhav.tiwari@tu-clausthal.de}
\author[1]{\fnm{Mohanad} \sur{Al-Ghobari}}\email{mohanad.al-ghobari@tu-clausthal.de}
\author*[1]{\fnm{Meng} \sur{Zhang}}\email{meng.zhang@tu-clausthal.de}
\author*[1]{\fnm{Andreas} \sur{Rausch}}\email{andreas.rausch@tu-clausthal.de}

\affil*[1]{\orgdiv{Institute for Software and Systems Engineering}, \orgname{Technical University of Clausthal}, \orgaddress{\city{Clausthal-Zellerfeld}, \country{Germany}}}

\abstract{
High-definition (HD) maps are core artifacts for automated driving systems, but their generation commonly relies on sensor-intensive mobile mapping campaigns, while quality assessment often depends on high-precision reference data. These dependencies make HD map engineering costly and difficult to apply in settings where specialised measurement data or independently measured reference maps are unavailable. This paper presents an engineering-oriented geo-data-driven workflow for HD map generation with integrated representation-level verification. The workflow uses openly available geo-engineering datasets as the primary input source and transforms them into lane-level HD map representations of existing road environments through explicit intermediate representations and processing stages. To assess the generated representations without external reference maps, the workflow integrates executable constraint-based verification into the engineering process. Selected constraints are derived from specifications relevant to automated driving and road-design guidelines. They are evaluated directly on the generated lanelet-based representation to detect geometric, topological, and elevation-related inconsistencies. The workflow is evaluated using real-world shapefile-based road-network data from four cities in Lower Saxony, Germany, and controlled defect-injection scenarios. The real-world evaluation shows that the generated map representations satisfy the selected constraints in the evaluated scenarios, while the defect-injection study demonstrates complete detection of the considered defect types without observed false positives. The results indicate that geo-data-driven HD map generation with integrated executable verification can provide a modular and inspectable complement to sensor-intensive mapping workflows under reduced sensing and reference-data availability.
}


\maketitle

\section{Introduction}\label{sec:introduction}

\subsection{Motivation}\label{sec:introduction-motivation}

High-definition (HD) maps are important artifacts in modern automated driving (AD) systems. They provide persistent prior knowledge about the geometry, topology and semantics of real-world road environments and complement onboard sensing, especially where local perception is limited by occlusion, range, weather, or computational constraints. As a result, HD maps are used across core modules such as localisation, perception, prediction, and motion planning. By providing a structured representation of road geometry, topology, and semantics, they help align these modules around a common model of the driving context. This makes HD maps safety-critical artifacts, because defects in map representations can propagate across modules and lead to system-level failures. For example, when a vehicle approaches a sharp curve after an uphill segment, the terrain may occlude the upcoming curvature from forward-facing cameras. In such cases, an HD map can support earlier speed adaptation through prior knowledge of the road geometry. However, inaccurate or inconsistent map information may instead lead to unsafe planning or control decisions.
From a software engineering perspective, the generation of HD maps for real-world road environments can be viewed as an engineering workflow that incrementally transforms information about physical road infrastructure into progressively refined digital map representations through multiple processing stages. Conventional HD map generation approaches exhibit three engineering characteristics commonly.

First,  conventional  HD  map generation approaches commonly rely on large-scale data collection using specialised high-end mapping sensor setups, such as 360 ° LiDAR, panoramic cameras, and high-precision GNSS/IMU units \cite{isprs-archives-XLIII-B4-2020-415-2020}. While such configurations provide rich and accurate mapping data, they require dedicated measurement campaigns and generate large raw-data volumes that involve substantial preprocessing, registration, fusion, and map-generation effort \cite{isprs-archives-XLIII-B4-2020-415-2020}. This sensing dependency motivates the investigation of alternative structured information sources for constructing initial HD map representations of existing road environments. Openly available geo-engineering data, such as road-network geometries, elevation data, and infrastructure-related attributes, describe existing physical road environments without requiring dedicated mapping campaigns. However, they are not HD maps: they may lack lane-level detail, explicit topology, complete semantics, and consistency guarantees required by automated driving systems.

Second, conventional approaches are typically implemented as tightly integrated processing pipelines. In multi-sensor fusion and SLAM-based approaches, map generation is commonly formulated as a joint estimation problem in which LiDAR, IMU, camera, and positioning data are optimised within a unified framework \cite{asrat2024comprehensive,bao2023review}. Although such approaches may contain intermediate data products, these representations are often pipeline-internal, tool-specific, or not exposed as independently verifiable engineering artifacts. This reduces transparency between processing stages, complicates localized diagnosis of transformation errors, and increases the risk that uncertainties propagate across subsequent stages. Similar challenges arise in many learning-based approaches, where intermediate processing is implicitly encoded in latent representations \cite{kwag2024review}, limiting interpretability and independent control of intermediate results. For geo-data-driven HD map generation, this motivates a modular workflow with explicit intermediate representations and inspectable transformation stages.

Third, map quality assessment is commonly tied to external reference data or indirect evaluation signals. Since HD maps do not have a true ground truth beyond the physical world itself, existing approaches approximate references using high-precision sensor data such as dense LiDAR point clouds or carefully collected mapping datasets. However, when workflows reduce reliance on raw sensor data,  representation-level consistency can no longer be directly assessed through comparison-based evaluation, increasing the risk that defects remain undetected. Together, these characteristics create a double dependency: map generation depends on specialized sensing campaigns, while quality assessment often depends on independently measured reference data. This work, therefore, investigates a modular and inspectable workflow for transforming open geo-engineering data into HD map representations of real-world road environments under reduced sensing dependency and integrates executable constraint-based verification for representation-level quality assessment.

\subsection{Challenges}\label{sec:introduction-problem}

Given the above observations, two engineering challenges arise.  


\textbf{Challenge 1: Designing a modular map generation pipeline with reduced sensing dependency}

The first challenge concerns how to generate HD map representations of existing real-world road environments without relying on specialized sensing campaigns as the primary data source. Addressing this challenge requires a modular and inspectable pipeline in which processing stages, intermediate representations, and transformation assumptions are made explicit.

\textbf{Challenge 2: Establishing representation-level quality assessment without measured references}

The second challenge concerns how to assess the consistency and structural quality of generated HD map representations when independently measured reference maps or high-precision ground-truth data are not available. Addressing this challenge requires practical geometric, topological, and semantic constraints that can be operationalized as executable verification checks directly on the generated map representation. Such checks should support representation-level defect detection rather than relying primarily on comparison with measured reference data or indirect downstream performance signals \cite{asrat2024comprehensive,kwag2024review}.

\subsection{Research Questions}\label{sec:introduction-rq}

To address the challenges outlined above, this work investigates the following research questions.

\begin{itemize}

\item RQ1: How can HD map representations be generated from open geo-engineering data through a modular pipeline that reduces dependence on specialized sensing campaigns while preserving inspectable intermediate artifacts?

\item RQ2: How can representation-level consistency of generated HD maps be assessed without relying on independently measured reference maps or high-precision reference data?
\begin{itemize}
\item RQ2.1: How can practical geometric, topological, and semantic constraints be defined for representation-level assessment of generated HD map representations?
\item RQ2.2: How can these constraints be operationalized as executable verification checks within the generation workflow?
\end{itemize}
\end{itemize}

\subsection{Paper Organization}

The remainder of this paper is organized as follows. Section 2 reviews related work on HD map generation and quality control. Section 3 presents the proposed modular HD map generation pipeline together with its integrated constraint-based verification approach. Section 4 describes the implementation of the pipeline. Section 5 evaluates the pipeline using real-world geo-data and controlled defect scenarios. Section 6 discusses the implications and limitations of the approach, and Section 7 concludes the paper and outlines future work. 

\section{Related Work}\label{sec:related-work}

\subsection{HD Map Generation Workflows}

HD map generation approaches can be broadly categorized into sensor-driven, learning-based, and geo-data-driven methods, depending on the primary data source and processing paradigm.





Sensor-driven HD map generation relies on large-scale data collection using mobile mapping systems equipped with LiDAR, cameras, and GNSS/IMU sensors. Such approaches typically involve multi-sensor fusion and tightly coupled processing stages \cite{bao2023review,asrat2024comprehensive}. Although they can achieve high geometric precision, they require dedicated measurement campaigns, large raw-data processing effort, and specialized sensor configurations, which limit modularity and scalability, and make intermediate representation difficult to inspect independently \cite{asrat2024comprehensive}.

Many recent approaches explore learning-based HD map generation from onboard perception data, often using Bird’s-Eye View (BEV) representations \cite{DBLP:journals/corr/abs-2107-06307,kwag2024review}. These methods support online inference in real-time and can reduce dependence on offline mapping campaigns. However, their intermediate reasoning is often encoded in implicit latent representations, which limits interpretability, traceability, and explicit structural diagnostics. Their performance also depends heavily on large annotated datasets and may degrade under distribution shifts \cite{kwag2024review}.

A third direction leverages structured geographic data, such as OpenStreetMap (OSM) or institutional GIS datasets, for HD map generation \cite{elghazaly2023high}. Such data provide broadly available information about road-network geometry, elevation, and infrastructure attributes, but they often lack lane-level detail, explicit topology, semantic completeness, and consistency guarantees required by AD systems \cite{elghazaly2023high}. Existing approaches address parts of this gap through structured transformation pipelines, lane-level map models, simulation-oriented conversion workflows, and OpenDRIVE-based road modelling \cite{6856487,lopez2018microscopic,DBLP:journals/corr/abs-1711-03938,asam_opendrive}. However, these approaches primarily focus on conversion and modelling, while systematic representation-level verification is typically external, limited, or not explicitly integrated into the generation workflow.

Overall, existing HD map generation approaches involve a trade-off between measurement dependency, automation, and workflow transparency. Sensor-driven methods provide high-precision map data but require specialized sensing campaigns; learning-based methods support automated or online generation but rely on learned latent representations and annotated datasets; geo-data-driven methods offer a more modular and inspectable basis but require explicit transformation and consistency checks. This motivates generation approaches that treat intermediate map representations as inspectable engineering artifacts rather than only as final outputs of a conversion pipeline. 

\subsection{HD Map Quality Control and Verification}

Ensuring representation-level consistency in HD maps is important since defects in geometry, topology, or semantics can affect downstream modules such as localization and motion planning. Existing HD map quality-control approaches differ primarily in how map quality is defined, what evidence is used, and whether diagnostics are available at the representation level. In the following, these approaches are broadly grouped into reference-based, learning-based, and constraint-based methods, with emphasis on their suitability for geo-data-driven methods without independently measured reference maps or high-precision reference data.

Reference-based approaches evaluate generated maps by comparing them with high-precision reference maps, dense point clouds, or annotated mapping datasets \cite{11376526,10614202,9888002}. They provide quantitative evidence of agreement with a measured reference representation, but depend on costly data collection and inherit the assumptions and limitations of the reference data. Downstream performance-based evaluation can provide additional end-to-end evidence, but it is indirect and often provides limited diagnostic insight into representation-level defects.

Learning-based approaches estimate map quality, uncertainty, or anomalies using machine learning techniques \cite{11147834,10.1007/978-3-658-45196-7_9}. They can handle noisy and incomplete inputs and may capture complex defect patterns, but their diagnostics are often model-dependent and less interpretable. Moreover, they do not necessarily provide explicit guarantees that geometric, topological, or semantic requirements are satisfied.

Constraint-based approaches define representation-level consistency through consistency constraints over geometric, topological, and semantic properties of the map representation \cite{10422044,qiu2021ontologybased,isprs-annals-X-1-W1-2023-621-2023}. They provide interpretable diagnostics without requiring direct comparison with measured high-precision reference maps or reference data. However, existing approaches often apply such checks as standalone verification steps rather than integrating them into the map generation workflow. They also provide limited support for structured defect analysis across multiple constraint categories and transformation stages.

In summary, existing quality-control approaches provide complementary evidence but remain insufficient for geo-data-driven HD map generation workflows. Reference-based methods require independently measured reference data, performance-based methods provide limited diagnostic localization, and learning-based methods often lack explicit structural guarantees. Constraint-based verification is well-suited for interpretable representation-level diagnostics but is commonly treated as a separate checking activity rather than as an integrated part of the generation workflow. This motivates a unified engineering workflow that combines geo-data-driven map generation with executable, representation-level constraint verification.

\section{Geo-data-Driven HD Map Generation Pipeline}\label{sec:method}
This section describes the proposed workflow architecture. The workflow is organized as a sequence of processing components that transform geospatial input data for a user-defined region of interest into a lane-level map representation and subsequently evaluate this representation through executable consistency checks. The description focuses on the workflow structure, component responsibilities, exchanged representations, and the integration point between map generation and verification.

\subsection{Initial Concept}\label{sec:problem}

The initial concept of the pipeline is derived from the two research questions addressed in this work. The first research question concerns the automated generation of lane-level HD maps from open geo-data, while the second concerns the verification of the generated maps with respect to geometric, topological, and semantic consistency constraints. This distinction motivates the separation of the workflow into a map generation part and a verification part. The generation part transforms open geo-data into a lane-level map representation, whereas the verification part evaluates the generated representation and reports detected inconsistencies.

Within the generation part, further decomposition is required because open geo-data are not directly suitable for HD map applications. They usually provide road geometries and attributes, but not complete lane-level structures, explicit lane boundaries, or verification-ready connectivity relations. Therefore, the system first retrieves and parses geo-data for a user-defined region of interest (ROI), then constructs a lane-level map representation, and finally converts it into a verification-compatible format. The ROI restricts processing to the relevant spatial area and avoids unnecessary extraction of large-scale geo-data. A frontend supports this workflow by allowing users to define the ROI, configure parameters, visualize the selected area, and initiate the workflow execution. These considerations lead to a system architecture, which is described in the following subsection.

\begin{figure}[t]
\centering
\includegraphics[width=\textwidth]{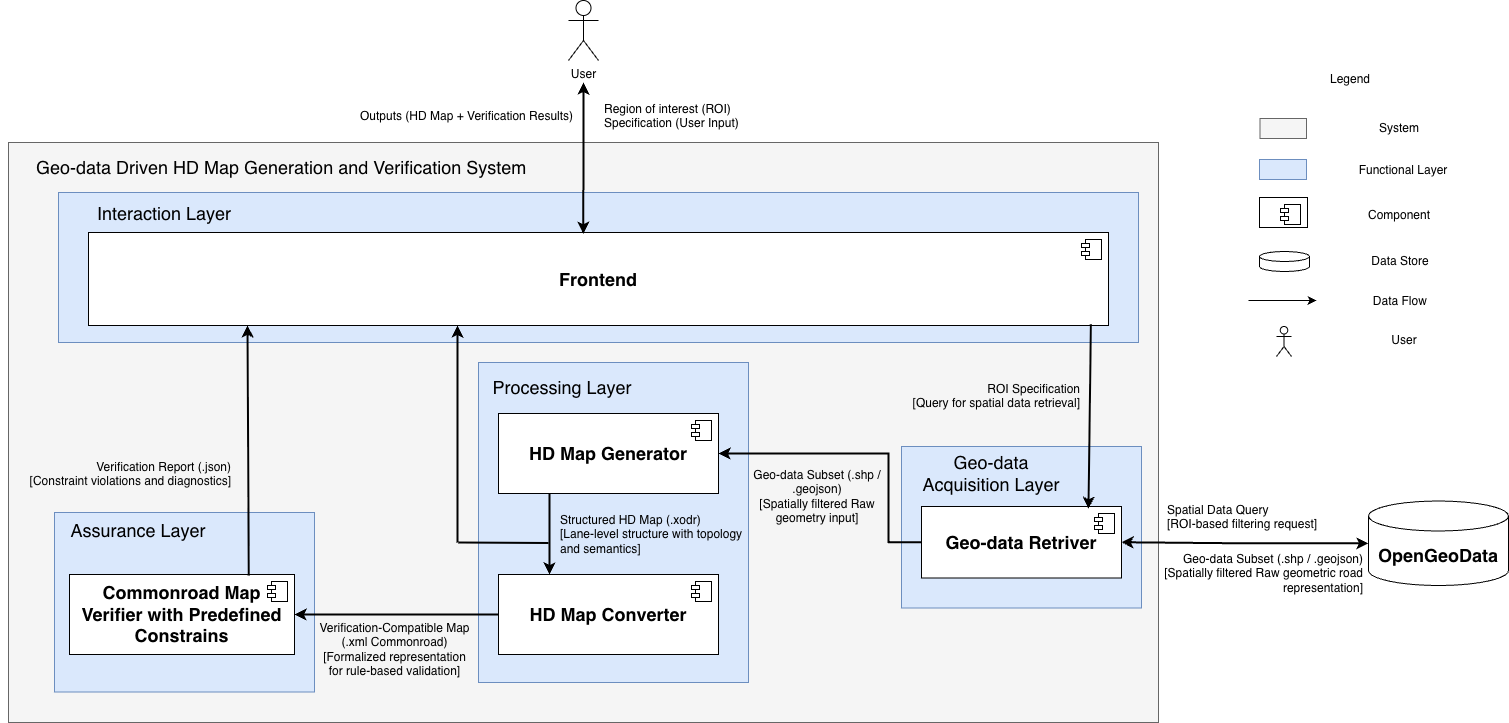}
\caption{System architecture of the geo-data-driven HD map generation and verification workflow.}
\label{fig:system-architecture}
\end{figure}

\subsection{System Overview}\label{sec:problem}

The system is organized as a modular workflow that separates user interaction, geo-data transformation, map representation conversion, and constraint-based verification. Fig. \ref{fig:system-architecture} illustrates the component-based architecture of the proposed geo-data–driven HD map generation and verification pipeline. The workflow starts from a user-defined ROI, retrieves the corresponding geo-data from available geospatial datasets, normalizes and transforms the retrieved data into a lane-level map representation, converts this representation into a verification-compatible format, and finally evaluates executable constraints to produce diagnostic reports describing detected violations and affected map elements.

The workflow is structured into three conceptual layers. The interaction layer defines the spatial and execution context of the pipeline. The transformation layer incrementally constructs the generated map representation from geo-data through explicit processing stages. The verification layer evaluates the generated representation against executable constraints and produces diagnostic feedback without modifying the map. As indicated in Fig. \ref{fig:system-architecture}, these conceptual layers are realised by five components with explicit interfaces: (1) the frontend component, (2) the geo-data retrieval and parsing component, (3) the HD map generator, (4) the HD map converter, and (5) the verification component.

\begin{figure}[H]
\centering
\includegraphics[width=\textwidth]{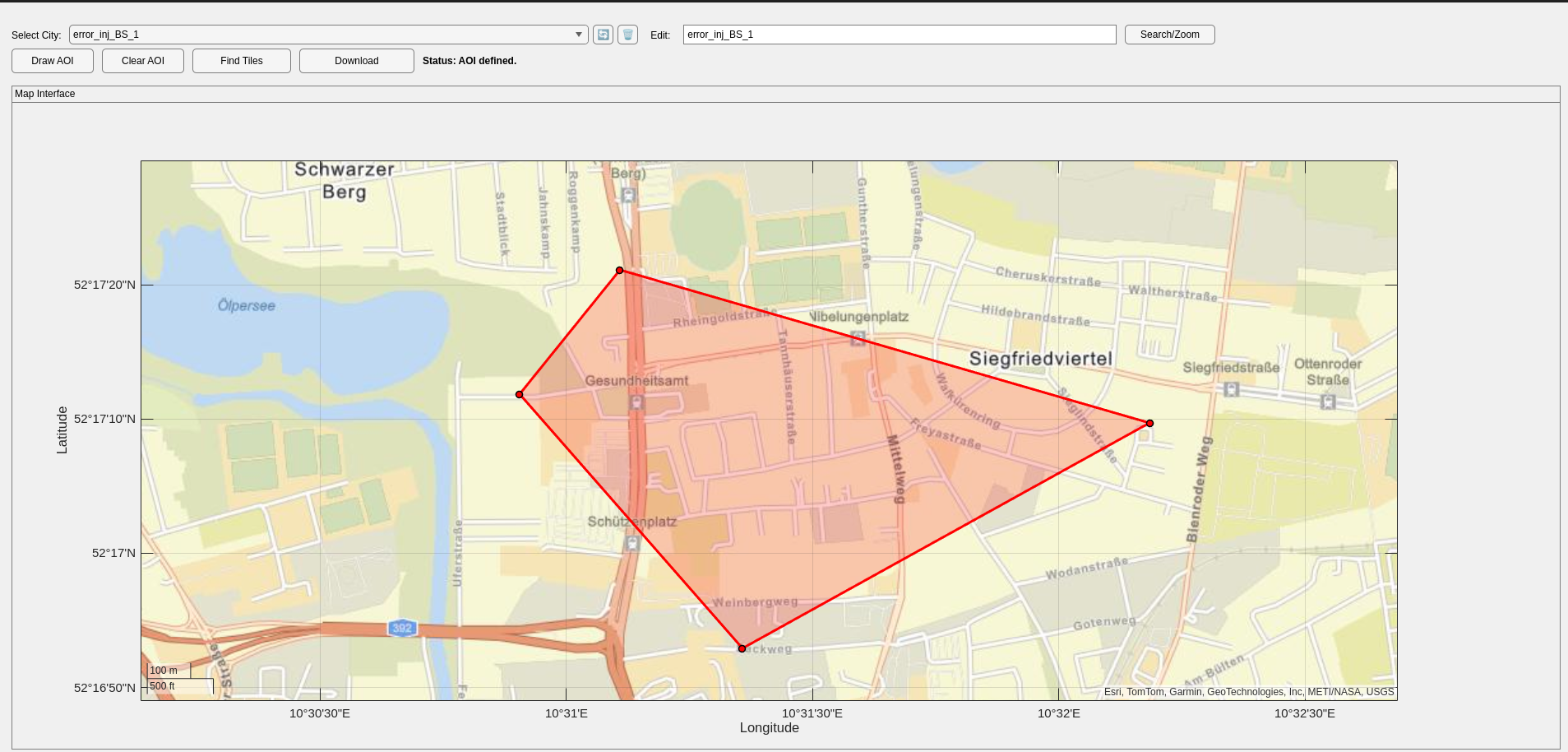}
\caption{Graphical selection of the region of interest (ROI) in the GUI. The target study area is defined interactively as a polygon over the basemap and is subsequently used for road-data clipping and DEM tile retrieval.}
\label{fig:aoi}
\end{figure}
The frontend component implements the interaction layer and provides the graphical interface for workflow orchestration, configuration management, and execution control. It captures the user-defined ROI and relevant configuration parameters and aggregates them into a structured execution context. In the current implementation, the ROI is defined interactively on a map view by selecting or drawing a spatial bounding region around the target road network. Internally, the ROI is represented as a bounding box in latitude–longitude coordinates and stored together with execution parameters such as the selected input dataset, coordinate reference settings, elevation source, export options, and verification configuration. Fig. \ref{fig:aoi} illustrates an example ROI selection in the frontend, where the user marks the target area before starting the generation workflow. The selected ROI is converted into spatial boundaries that are used consistently for geo-data extraction and elevation sampling. By managing ROI selection, parameter handling, and initialisation of downstream processing stages, the frontend ensures reproducible and consistent pipeline execution.

The geo-data retrieval and parsing component constitutes the first stage of the transformation layer. Its purpose is to extract the road information required for a target study area from raw geospatial data and convert it into structured inputs for subsequent modelling. In the current implementation, the input geo-data are obtained from official open geospatial datasets published by the state of Lower Saxony, Germany, including Basis-DLM - Shapefile~\cite{lgln_basis_dlm_shapefile} and Digitales Geländemodell (DGM1)~\cite{lgln_dgm1}. Basis-DLM provides vector-based representations of surface objects such as roads, water bodies, vegetation, and settlement areas, and is used here as the primary source of road centerlines and road attributes. As part of the ATKIS topographic data framework, it provides geometrically consistent and attribute-rich road data in the ETRS89 / UTM zone 32N reference system. DGM1 is a bare-earth digital terrain model derived from airborne laser scanning and provided in \mbox{$1 \times 1$ km} tiles with a raster resolution of \mbox{$1$ m}, making it suitable for fine-grained elevation sampling in lane-level HD map generation. Using the ROI boundary as a spatial constraint, the component filters road objects from the global road layer, referred to here as the VER1 layer, to produce a local road dataset. It further parses the key road attributes and organises them into semantically relevant descriptors, primarily including \texttt{OBJID}, \texttt{OBJART}, \texttt{OBJART\_TXT}, \texttt{NAM}, \texttt{BEZ}, \texttt{FSZ}, \texttt{BRF}, \texttt{FAR}, \texttt{FKT}, \texttt{WDM}, and \texttt{ZUS}. Among these, \texttt{FSZ}, \texttt{BRF}, and \texttt{FAR} provide the core basis for recovering lane count, road width, and traffic direction semantics. In addition, the component performs coordinate reference system validation, projection information export, and caching of locally clipped results, thereby providing stable and standardised inputs for the downstream mapping stage.

The HD map generator converts the resulting local road centerlines and their key attributes into a lane-level RoadRunner HD Map. It first transforms the input centerlines into a local coordinate system in order to reduce the numerical scale effects of geodetic coordinates on subsequent geometric computation. It then clusters centerline endpoints to identify topological nodes in the road network and projects nodes located within road interiors back onto the centerlines to split the roads into segment units suitable for lane-level modelling. During semantic recovery, the system uses \texttt{FSZ} to infer the number of lanes, \texttt{BRF} to determine the lateral road width and lane expansion scale, and \texttt{FAR} to determine one-way or two-way traffic properties and assign lane travel directions accordingly. Based on these inferred semantics, the system automatically generates lane centerlines and lane boundaries and organizes them into \textit{lane}, \textit{lane boundary}, and \textit{lane group} objects in the RoadRunner HD Map. For topology construction, it establishes predecessor and successor relations from the spatial clustering of lane endpoints and recovers basic traffic connectivity using direction constraints, distance constraints, and a greedy one-to-one matching strategy. In intersection areas, candidate connections are further classified according to the heading relationships between nearby incoming lanes and outgoing lanes, allowing straight, left-turn, and right-turn movements to be added and thereby enriching the topological semantics of the local road network. For three-dimensional map generation, DEM data are introduced after the two-dimensional lane geometry has been constructed. By selecting, mosaicking, and sampling elevation tiles covering the ROI, and by applying missing-value filling, smoothing, and grade constraints, elevation information is propagated to both lane centerlines and boundaries, producing an RRHD map with full 3D coordinates. Finally, the system outputs a local-coordinate \texttt{.rrhd} file and restores its corresponding global-coordinate version using the saved origin offset. The resulting map is then imported into RoadRunner for automatic scene construction and exported as \texttt{.rrscene}, \texttt{.fbx}, and \texttt{.xodr} files for downstream.

The HD map converter \cite{Althoff2018b} transforms OpenDRIVE road network descriptions into CommonRoad scenario representations. The method converts roads and lane structures defined by reference lines and lateral parameters in OpenDRIVE into an explicit lanelet-based road network in CommonRoad. In OpenDRIVE, road geometry is defined by reference lines, while lane boundaries are implicitly described through lateral offsets, width functions, and lane sections. In CommonRoad, roads are represented as lanelets with explicit left boundaries, right boundaries, and centerlines, together with predecessor, successor, and lateral adjacency relations. The conversion therefore extracts the reference lines, lane sections, width functions, elevation information, and lateral profile features from the OpenDRIVE file, reconstructs the inner and outer lane boundaries along the longitudinal path parameter, and discretely samples them to generate lanelet boundary point sequences. The centerline of each lanelet is computed from the corresponding boundary midpoints, and the connectivity information from the original road description is used to establish the lanelet topology. For three-dimensional road structures, longitudinal elevation and lateral profile information are combined with the recovered planar boundaries to preserve height-related geometry. The implementation is based on Python, uses \texttt{commonroad-io} to construct CommonRoad \texttt{Scenario}, \texttt{LaneletNetwork}, and \texttt{Lanelet} objects, and employs \texttt{pyproj} for coordinate transformation and georeferencing. The resulting road network is exported as a standard CommonRoad XML scenario file.

The converted CommonRoad map is verified through a rule-based process to assess road-structure consistency. The process takes the Lanelet network as input and evaluates predefined validity rules over lanelet boundary geometry, predecessor-successor relations, lateral adjacency relations, and road-element references. The rules are expressed in higher-order logic, with their syntax defined and parsed by ANTLR \cite{antlr4_github} to obtain executable expressions. CommonRoad road elements are then mapped to logical domains and evaluated through predicate and term functions. When a rule is violated, the corresponding lanelet or road-element identifiers are recorded as invalid states, providing explicit locations of structural inconsistencies. For large road networks, the verification can be applied to partitioned sub-networks and executed in parallel to improve efficiency.

To support rule extension, the implementation further includes an LLM-assisted rule generation script \cite{he2025llmassistedtooljointgeneration}. Given a natural-language rule requirement, the script constructs a prompt containing the current code context and invokes a large language model to generate modifications such as rule identifiers, formula mappings, and predicate functions. The generated outputs are integrated into the existing verification framework through structured patches or operations and checked by a basic smoke test before being used in the higher-order-logic verification workflow. Thus, the LLM is used to assist rule development and code generation, while the final map verification remains a deterministic higher-order-logic evaluation process. The overall workflow is implemented in Python, operates directly on \texttt{commonroad-io} map objects, and organises the results as rule-wise invalid states for subsequent analysis or text report generation.


\begin{figure}[H]
\centering
\includegraphics[width=\textwidth]{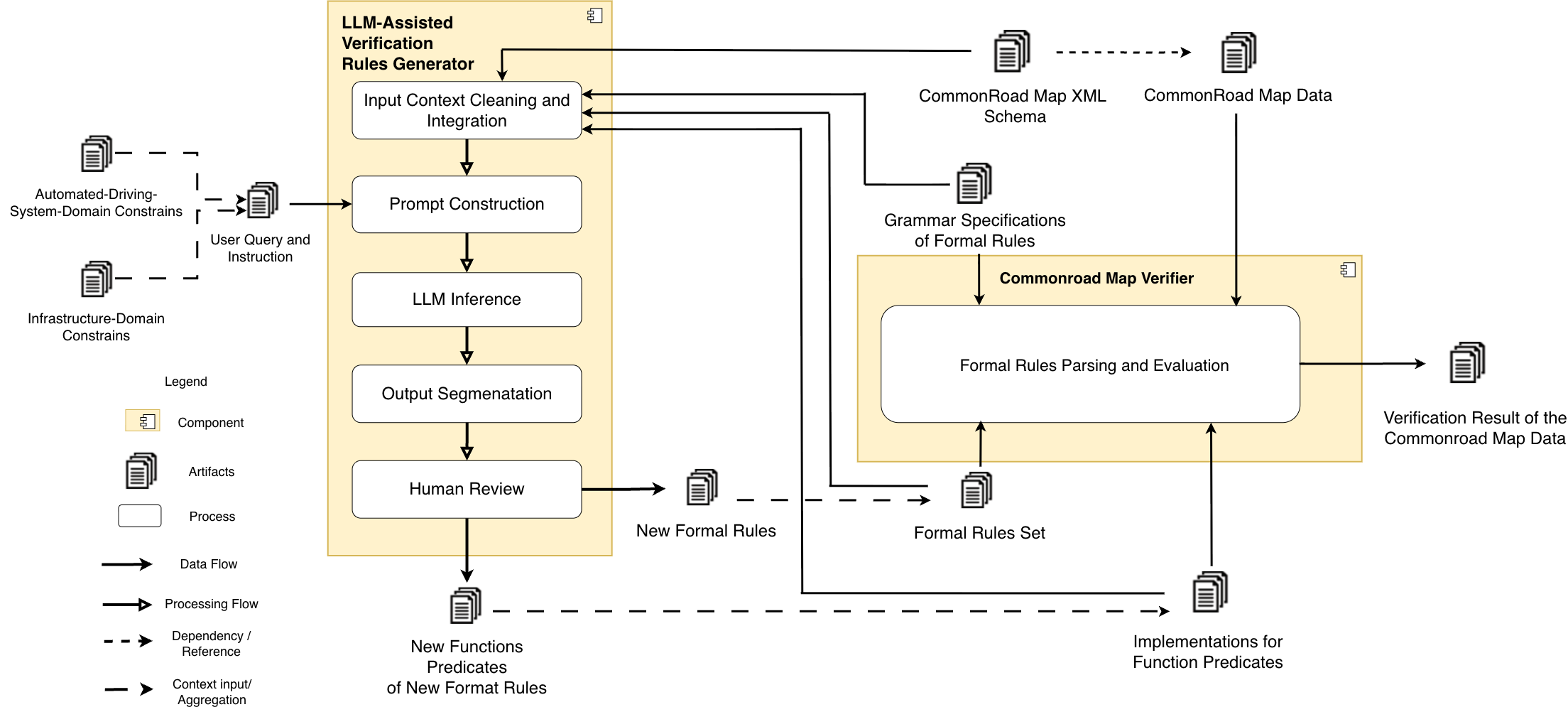}
\caption{LLM-assisted map verification workflow \cite{he2025llmassistedtooljointgeneration}. Constraint rules are constructed from heterogeneous sources with LLM assistance and then evaluated over the map representation by the verification component, producing interpretable diagnostics.}
\label{fig:verification-execution-verifier}
\end{figure}

\section{Evaluation}\label{sec:evaluation}

\subsection{Evaluation Setup}

The evaluation follows a sequential procedure in which selected regions of interest (ROIs) are used to generate HD maps, and the corresponding verification reports are manually inspected to assess whether the verification component identifies meaningful inconsistencies and provides interpretable diagnostic information. The evaluation comprises three complementary case studies: (1) verification of automated-driving domain constraints on real-world generated maps, (2) verification of road-design constraints, and (3) controlled defect injection. Across all studies, the verification pipeline processes the generated lanelet networks, evaluates the selected constraints, and produces reports containing detected violations and their spatial locations. The reported violations are then manually compared with the corresponding raw geo-data and generated map artifacts to assess whether they indicate actual inconsistencies. The evaluation is intentionally scoped as a feasibility-oriented engineering study rather than a comprehensive benchmarking campaign.

\subsubsection{Case Study 1: Autonomous Driving Domain Constraints on Real-World Maps.}
The first case study evaluates representation-level map constraints relevant to automated driving. For this purpose, 45 HD map samples are generated from real-world shapefile-based geospatial road-network data across four cities in  Lower  Saxony,  Germany:  Hannover, Braunschweig, Goslar, and Hildesheim. The selected regions of interest cover diverse urban road scenarios, including straight roads, curved roads, T-junctions, intersections, and merging lanes. The evaluated constraints address common consistency requirements for automated-driving map representations \cite{autoware_system}, including elevation completeness, topological consistency, and geometric validity of lane boundaries.

Three constraint categories are evaluated:
\begin{itemize}
    \item \textbf{Elevation Completeness:}
    Validates that all boundary points contain finite z-coordinates.
    \item \textbf{Successor Non-Self-Loop:}
    Ensures that a lanelet does not reference itself as a successor.
    \item \textbf{Polyline Validity:}
    Requires each boundary polyline to contain at least two geometrically distinct points.
\end{itemize}

\subsubsection{Case Study 2: Road Design Constraints.}

The second case study evaluates whether the map quality assessment in the pipeline can also be applied to constraints derived from road-design guidelines. For this purpose, a constraint is formulated based on the RASt 06 road-design guideline \cite{rast06}. The evaluation focuses on built-up urban areas, since the selected requirement is applicable to this road context and can be assessed using the geometric information available in the generated lanelet networks. A total of 13 HD maps are generated across the same four cities used in Case Study 1, covering comparable urban road scenarios. Verification is performed using the same batch-processing pipeline as in the first case study. The evaluated constraint is defined as follows:  

\begin{itemize}
    \item \textbf{Minimum Turning Radius:}
    Validates that the turning radius of each lanelet remains above $10.0\,\mathrm{m}$, as specified by RASt~06 for built-up urban roads.
\end{itemize}

\subsubsection{Case Study 3: Controlled Defect Injection.}

The first two case studies assess the quality of generated maps by applying selected constraints to real-world map samples and manually inspecting the reported violations. However, since naturally occurring defects are not systematically controlled, these studies cannot fully assess whether the verification rules detect known defect types reliably. Therefore, a third case study is designed to evaluate defect detection under controlled conditions.
The third case study uses controlled defect injection to systematically assess detection behaviour and completeness. Unlike the previous case studies, where violations may occur naturally, known defects are intentionally introduced into a representative urban map from Braunschweig, Germany. Multiple modified map versions are generated from this baseline, each containing predefined defects. Three defect categories are considered, each with 10 injected defects:

\begin{itemize}

    \item \textbf{Elevation Completeness Defect:}
    Non-finite z-coordinates (NaN) are introduced into selected lanelets to simulate incomplete elevation data.

    \item \textbf{Lane Width Defect:}
    The lateral distance between lane boundaries is artificially reduced to create unrealistically narrow lane sections.

    \item \textbf{Successor Self-Loop Defect:}
    Self-loop successor relations are introduced into selected lanelets to create invalid topological cycles.

\end{itemize}

\subsection{Metrics}

Case Studies 1 and 2 do not provide complete ground-truth annotations that distinguish all consistent and inconsistent map elements. Therefore, precision- and recall-based metrics cannot be computed reliably for these real-world studies, since the complete set of true violations and non-violations is unknown. Following prior work on constraint-based and ontology-driven validation \cite{qiu2021ontologybased,10422044,RDFvalidation}, the evaluation instead reports detected violations per map and per constraint category, followed by manual inspection of the reported diagnostics.
In contrast, Case Study 3 provides ground truth by map generation. Since the injected defects and their locations are known a priori, the controlled defect injection study enables quantitative assessment of detection behaviour, following metrics commonly applied in map change detection and anomaly detection \cite{s21072477,11147834}. For this study, precision is used to measure how many reported violations correspond to injected defects:

\begin{equation}
\text{Precision} = \frac{N_{\text{correctly detected}}}{N_{\text{reported}}}
\end{equation}
where $N_{\text{reported}}$ denotes the number of reported violations and $N_{\text{correctly detected}}$ the number corresponding to injected defects.

Detection completeness is evaluated using the true positive rate (TPR):
\begin{equation}
\text{TPR} = \frac{N_{\text{correctly detected}}}{N_{\text{injected}}}
\end{equation}
where $N_{\text{injected}}$ denotes the total number of injected defects.

In addition, \textbf{false positives (FP)} are analyzed as reported violations that do not correspond to injected defects. This distinction allows the evaluation to assess both the correctness of reported diagnostics and the completeness of defect detection under controlled conditions.

\subsection{Results}

\subsubsection{Case Study 1 and 2: Real-World Maps.}

The evaluation demonstrates the feasibility of the proposed engineering-oriented workflow for constructing and assessing HD maps from real-world geo-data. As summarized in Table~\ref{tab:real_world_case_studies_results}, no violations were detected across the 45 maps evaluated in Case Study~1 for elevation, polyline, and self-loop constraints, nor across the 13 urban road maps evaluated in Case Study~2 for the minimum turning radius constraint. These results indicate that the proposed workflow can generate structurally consistent HD maps across diverse urban scenarios using geo-data as the primary input source. In addition, the constraint-based verification component successfully detected controlled defects with high diagnostic precision.

These results indicate that the generated maps satisfy the selected representation-level and road-design constraints within the evaluated scenarios. They also demonstrate that the proposed workflow can process real-world shapefile-based geo-data, generate lanelet-based map representations, and execute constraint-based verification in a batch-processing pipeline. However, since no complete ground-truth annotations are available for these real-world maps, the results should be interpreted as evidence of feasibility and consistency with respect to the selected constraints, rather than as a comprehensive proof of map correctness.
\subsubsection{Case Study 3: Defect Injection.}

Table~\ref{tab:injected_error_case_study_results} summarizes the results of the controlled defect injection experiments. No violations were detected in the original baseline map before defect injection. After injecting defects, the verifier detected all 10 elevation completeness defects, all 10 lane-width defects, and all 10 successor self-loop defects. No false positives were reported for any of the evaluated defect categories.
Consequently, precision and true positive rate (TPR) both reached 1.00 for all three defect categories. These results show that the implemented constraints can reliably detect the considered defect types under controlled conditions. In combination with the real-world evaluations in Case Studies 1 and 2, the defect injection study (Case Study 3) provides targeted evidence that the verification component can correctly and completely detect the selected defect types under controlled conditions.
\begin{table}[t]
\centering
\caption{Verification results for the real-world map evaluation in Case Studies~1 and~2, showing the number of detected violations per evaluated constraint.}
\label{tab:real_world_case_studies_results}
\small
\begin{tabular}{@{}p{2.4cm}p{5.6cm}cc@{}}
\toprule
\textbf{Case Study} & \textbf{Constraint} & \textbf{Maps} & \textbf{Violations Reported} \\
\midrule
\multirow{3}{*}{Case Study 1} & Elevation Completeness & 45 & 0 \\
& Successor Non-Self-Loop   & 45 & 0 \\
& Polyline Validity & 45 & 0 \\
\midrule
Case Study 2 & Minimum Turning Radius & 13 & 0 \\
\bottomrule
\end{tabular}
\end{table}

\begin{table}[t]
\centering
\caption{Detection performance for the controlled defect injection experiments in Case Study~3. ``Base.'' denotes violations detected in the original baseline map, ``Inj.'' the number of injected defects, and ``Det.'' the number of detected defects reported by the verifier.}
\label{tab:injected_error_case_study_results}
\small
\begin{tabular}{@{}p{4.0cm}ccccccc@{}}
\toprule
\textbf{Constraint} & \textbf{Base.} & \textbf{Inj.} & \textbf{Det.} & \textbf{TP} & \textbf{FP} & \textbf{Precision} & \textbf{TPR} \\
\midrule
Elevation  & 0 & 10 & 10 & 10 & 0 & 1.000 & 1.000 \\
Lane Width  & 0 & 10 & 10 & 10 & 0 & 1.000 & 1.000 \\
Self-Loop Connections & 0 & 10 & 10 & 10 & 0 & 1.000 & 1.000 \\
\bottomrule
\end{tabular}
\end{table}

\section{Discussion}\label{sec:discussion}
This work investigated how HD map generation and representation-level quality assessment can be integrated into a modular engineering workflow.  The proposed approach combines geo-data–driven map generation with executable constraint-based verification and organizes both activities within an inspectable workflow with explicit intermediate representations.

The workflow is not intended to achieve centimeter-level geometric accuracy comparable to dedicated sensor-intensive mapping systems. Since the generated maps depend on the completeness, resolution, and attribute quality of publicly available geo-data, their absolute geometric accuracy is inherently bounded by the source data. The primary objective is therefore not precise physical alignment with the real-world environment but the generation of lane-level HD map representations satisfying selected structural constraints with integrated verification support.

The results indicate that structured geo-data can provide a practical basis for lane-level HD map generation without requiring tightly coupled sensor-processing pipelines or specialized mapping systems. In contrast to traditional SLAM-based approaches, the proposed workflow preserves explicit intermediate representations and separates data retrieval, map generation, representation conversion, and verification into distinct processing stages. This improves inspectability, supports incremental debugging, and allows individual components or constraints to be extended independently. 

Map quality is assessed through executable consistency constraints evaluated directly on the generated map representations rather than through comparison against external reference maps. The evaluated constraints capture baseline geometric and topological consistency conditions derived from automated driving system specifications and infrastructure-oriented road-design rules. The real-world case studies show that the generated map representations satisfy the selected constraints in the evaluated scenarios, while the controlled defect injection study demonstrates that the implemented rules can detect the considered defect types with interpretable diagnostics under controlled conditions. These findings support the feasibility of integrating constraint-based verification into a geo-data–driven HD map generation workflow.
Several limitations remain. First, the workflow depends on the completeness and consistency of geo-data attributes such as lane count, road direction, and road class. Missing or ambiguous attributes require heuristic assumptions, which may affect the correctness of the generated lane-level map representation. Second, the current constraint set primarily targets baseline geometric and topological consistency conditions and should be regarded as an initial feasibility-oriented rule set rather than a comprehensive map-quality model. Third, verification outcomes may be influenced by representation conversion effects, since constraints are evaluated on converted CommonRoad representations rather than directly on the originally generated OpenDRIVE maps. Finally, the evaluation is limited to selected urban scenarios in Lower Saxony, Germany, and primarily reflects European road-design assumptions, including those derived from RASt 06 \cite{rast06}. 
Future work should extend the constraint set toward richer semantic and behavioral verification, including traffic-rule consistency, signal and intersection semantics, lane-transition legality, and route-level connectivity. Additional data sources, such as aerial imagery, point clouds, and onboard sensor observations may further improve semantic reconstruction and robustness. Broader evaluations across additional geographic regions, road structures, and defect categories are also required to assess the general applicability and robustness of the proposed workflow.

\section{Conclusion}\label{sec:conclusion}

This work presented a geo-data–driven workflow for HD map generation with integrated representation-level quality assessment. The approach combines modular map generation with executable constraint-based verification, enabling direct and interpretable consistency analysis of generated map representations without relying on external reference maps or data.

The evaluation demonstrates the feasibility of generating lane-level HD map representations from publicly available geo-data and operationalizing geometric and topological consistency checks as executable verification rules. The real-world case studies show that the generated map representations satisfy the selected constraints in the evaluated scenarios, while the controlled defect injection study confirms that the implemented rules can detect the considered defect types with interpretable diagnostics. The results suggest that integrated verification can provide a practical and inspectable complement to traditional sensor-intensive mapping workflows, particularly for early-stage map generation, prototyping, and consistency-oriented assessment.

Additionally, the current approach remains limited by the quality and completeness of the underlying geo-data, representation conversion effects, and the restricted evaluation scope. Future work should extend the constraint set toward verification for richer semantics, incorporate additional data sources, and evaluate the workflow across broader geographic and road-design contexts.

\bibliography{sn-bibliography}

@article{asrat2024comprehensive,
  title={A comprehensive survey on high-definition map generation and maintenance},
  author={Asrat, Kaleab Taye and Cho, Hyung-Ju},
  journal={ISPRS International Journal of Geo-Information},
  volume={13},
  number={7},
  pages={232},
  year={2024},
  publisher={MDPI}
}

@article{bao2023review,
  title={A review of high-definition map creation methods for autonomous driving},
  author={Bao, Zhibin and Hossain, Sabir and Lang, Haoxiang and Lin, Xianke},
  journal={Engineering Applications of Artificial Intelligence},
  volume={122},
  pages={106125},
  year={2023},
  publisher={Elsevier}
}

@article{kwag2024review,
  title={A Review on End-to-End High-Definition Map Generation},
  author={Kwag, Jiyong and Toth, Charles},
  journal={The International Archives of the Photogrammetry, Remote Sensing and Spatial Information Sciences},
  volume={48},
  pages={187--194},
  year={2024},
  publisher={Copernicus Publications G{\"o}ttingen, Germany}
}

@article{elghazaly2023high,
  title={High-definition maps: Comprehensive survey, challenges, and future perspectives},
  author={Elghazaly, Gamal and Frank, Rapha{\"e}l and Harvey, Scott and Safko, Stefan},
  journal={IEEE Open Journal of Intelligent Transportation Systems},
  volume={4},
  pages={527--550},
  year={2023},
  publisher={IEEE}
}

@INPROCEEDINGS{11376526,
  author={Kalenda, Marie-Ngoïe Badibanga and Bonnifait, Philippe and Mittet, Marie-Anne},
  booktitle={2025 IEEE International Conference on Vehicular Electronics and Safety (ICVES)}, 
  title={Vector Map Quality Metrics for Contextual Autonomous Driving Systems}, 
  year={2025},
  volume={},
  number={},
  pages={367-373},
  doi={10.1109/ICVES65691.2025.11376526}}

@ARTICLE{10614202,
  author={Chiang, Kai-Wei and Srinara, Surachet and Chiu, Yu-Ting and Tsai, Syun and Tsai, Meng-Lun and Satirapod, Chalermchon and El-Sheimy, Naser and Ai, Mengchi},
  journal={IEEE Internet of Things Journal}, 
  title={Creation and Verification of High-Definition Point Cloud Maps for Autonomous Vehicle Navigation}, 
  year={2024},
  volume={11},
  number={23},
  pages={37582-37598},
  doi={10.1109/JIOT.2024.3435344}}

@INPROCEEDINGS{10422044,
  author={Maierhofer, Sebastian and Ballnath, Yannick and Althoff, Matthias},
  booktitle={2023 IEEE 26th International Conference on Intelligent Transportation Systems (ITSC)}, 
  title={Map Verification and Repairing Using Formalized Map Specifications}, 
  year={2023},
  volume={},
  number={},
  pages={1277-1284},
  doi={10.1109/ITSC57777.2023.10422044}}

@inproceedings{
qiu2021ontologybased,
title={Ontology-Based Map Data Quality Assurance},
author={Haonan Qiu and Adel Ayara and Birte Glimm},
booktitle={Eighteenth Extended Semantic Web Conference - Research Track},
year={2021},
url={https://openreview.net/forum?id=xNHyHnpLT5S}
}

@Article{isprs-annals-X-1-W1-2023-621-2023,
AUTHOR = {Chiang, K.-W. and Tsai, M.-L. and Lin, S. and Huang, Y.-E. and Zeng, J.-C. and Chang, Y.-F. and Chen, J.-A. and Huang, Y.-C. and Yang, C.-S. and Juang, J.-C. and Wang, C.-K. and Lin, C.-F. and Lee, J. and Darweesh, H. and Li, P.-L.},
TITLE = {ESTABLISHMENT OF HD MAPS VERIFICATION AND VALIDATION PROCEDURE WITH OPENDRIVE AND AUTOWARE (LANELET2) FORMATS},
JOURNAL = {ISPRS Annals of the Photogrammetry, Remote Sensing and Spatial Information Sciences},
VOLUME = {X-1/W1-2023},
YEAR = {2023},
PAGES = {621--627},
URL = {https://isprs-annals.copernicus.org/articles/X-1-W1-2023/621/2023/},
DOI = {10.5194/isprs-annals-X-1-W1-2023-621-2023}
}

@Article{isprs-archives-XLIII-B4-2020-415-2020,
AUTHOR = {Tsushima, F. and Kishimoto, N. and Okada, Y. and Che, W.},
TITLE = {CREATION OF HIGH DEFINITION MAP FOR AUTONOMOUS DRIVING},
JOURNAL = {The International Archives of the Photogrammetry, Remote Sensing and Spatial Information Sciences},
VOLUME = {XLIII-B4-2020},
YEAR = {2020},
PAGES = {415--420},
URL = {https://isprs-archives.copernicus.org/articles/XLIII-B4-2020/415/2020/},
DOI = {10.5194/isprs-archives-XLIII-B4-2020-415-2020}
}

@ARTICLE{9888002,
  author={Chiang, Kai-Wei and Zeng, Jhih-Cing and Tsai, Meng-Lun and Darweesh, Hatem and Chen, Pin-Xu and Wang, Chi-Kuei},
  journal={IEEE Journal of Selected Topics in Applied Earth Observations and Remote Sensing}, 
  title={Bending the Curve of HD Maps Production for Autonomous Vehicle Applications in Taiwan}, 
  year={2022},
  volume={15},
  number={},
  pages={8346-8359},
  doi={10.1109/JSTARS.2022.3204306}}

@INPROCEEDINGS{11147834,
  author={Shaw, Ankit Kumar and Jiang, Kun and Wen, Tuopu and Sah, Chandan Kumar and Shi, Yining and Yang, Mengmeng and Yang, Diange and Lian, Xiaoli},
  booktitle={2025 IEEE/CVF Conference on Computer Vision and Pattern Recognition Workshops (CVPRW)}, 
  title={CleanMAP: Distilling Multimodal LLMs for Confidence-Driven Crowdsourced HD Map Updates}, 
  year={2025},
  volume={},
  number={},
  pages={3798-3807},
  doi={10.1109/CVPRW67362.2025.00365}}

@InProceedings{10.1007/978-3-658-45196-7_9,
author="Pfeifle, Martin
and Glander, Karl-Heinz
and Sz{\'a}nt{\'o}, Marcell
and Pfeifle, Timo",
editor="Heintzel, Alexander",
title="Enhancing Digital Maps with AI-Derived Confidence Information",
booktitle="Automatisiertes Fahren 2024",
year="2024",
publisher="Springer Fachmedien Wiesbaden",
address="Wiesbaden",
pages="105--117",
isbn="978-3-658-45196-7"
}

@article{DBLP:journals/corr/abs-2107-06307,
  author       = {Qi Li and
                  Yue Wang and
                  Yilun Wang and
                  Hang Zhao},
  title        = {HDMapNet: An Online {HD} Map Construction and Evaluation Framework},
  journal      = {CoRR},
  volume       = {abs/2107.06307},
  year         = {2021},
  url          = {https://arxiv.org/abs/2107.06307},
  eprinttype   = {arXiv},
  eprint       = {2107.06307},
  bibsource    = {dblp computer science bibliography, https://dblp.org}
}

@INPROCEEDINGS{6856487,
  author={Bender, Philipp and Ziegler, Julius and Stiller, Christoph},
  booktitle={2014 IEEE Intelligent Vehicles Symposium Proceedings}, 
  title={Lanelets: Efficient map representation for autonomous driving}, 
  year={2014},
  volume={},
  number={},
  pages={420-425},
  doi={10.1109/IVS.2014.6856487}}

@inproceedings{lopez2018microscopic,
  title={Microscopic traffic simulation using sumo},
  author={Lopez, Pablo Alvarez and Behrisch, Michael and Bieker-Walz, Laura and Erdmann, Jakob and Fl{\"o}tter{\"o}d, Yun-Pang and Hilbrich, Robert and L{\"u}cken, Leonhard and Rummel, Johannes and Wagner, Peter and Wie{\ss}ner, Evamarie},
  booktitle={2018 21st international conference on intelligent transportation systems (ITSC)},
  pages={2575--2582},
  year={2018},
  organization={Ieee}
}

@article{DBLP:journals/corr/abs-1711-03938,
  author       = {Alexey Dosovitskiy and
                  Germ{\'{a}}n Ros and
                  Felipe Codevilla and
                  Antonio M. L{\'{o}}pez and
                  Vladlen Koltun},
  title        = {{CARLA:} An Open Urban Driving Simulator},
  journal      = {CoRR},
  volume       = {abs/1711.03938},
  year         = {2017},
  url          = {http://arxiv.org/abs/1711.03938},
  eprinttype   = {arXiv},
  eprint       = {1711.03938},
  bibsource    = {dblp computer science bibliography, https://dblp.org}
}

@misc{he2025llmassistedtooljointgeneration,
      title={LLM-Assisted Tool for Joint Generation of Formulas and Functions in Rule-Based Verification of Map Transformations}, 
      author={Ruidi He and Yu Zhang and Meng Zhang and Andreas Rausch},
      year={2025},
      eprint={2511.01423},
      archivePrefix={arXiv},
      primaryClass={cs.SE},
      url={https://arxiv.org/abs/2511.01423}, 
}

@misc{asam_opendrive,
  author       = {{ASAM e.V.}},
  title        = {{OpenDRIVE Format Specification}},
  year         = {2023},
  note         = {Version 1.7},
  url          = {https://www.asam.net/standards/detail/opendrive/}
}

@misc{autoware_system,
  author       = {{Autoware Foundation}},
  title        = {Autoware: Open-source software for autonomous driving},
  year         = {2023},
  howpublished = {\url{https://www.autoware.org/}},
  note         = {Accessed: 2026-04}
}

@book{rast06,
  author    = {{FGSV}},
  title     = {Richtlinien für die Anlage von Stadtstraßen (RASt 06)},
  year      = {2006},
  publisher = {Forschungsgesellschaft für Straßen- und Verkehrswesen},
  address   = {Köln, Germany}
}

@article{RDFvalidation,
author = {De Meester, Ben and Heyvaert, Pieter and Arndt, Dörthe and Dimou, Anastasia and Verborgh, Ruben},
year = {2020},
month = {11},
pages = {117-142},
title = {RDF graph validation using rule-based reasoning},
volume = {12},
journal = {Semantic Web},
doi = {10.3233/SW-200384}
}

@Article{s21072477,
AUTHOR = {Zhang, Pan and Zhang, Mingming and Liu, Jingnan},
TITLE = {Real-Time HD Map Change Detection for Crowdsourcing Update Based on Mid-to-High-End Sensors},
JOURNAL = {Sensors},
VOLUME = {21},
YEAR = {2021},
NUMBER = {7},
ARTICLE-NUMBER = {2477},
URL = {https://www.mdpi.com/1424-8220/21/7/2477},
PubMedID = {33918443},
ISSN = {1424-8220},
DOI = {10.3390/s21072477}
}

@inproceedings{Althoff2018b,

	author = {Matthias Althoff and Stefan Urban and Markus Koschi},
	title = {Automatic Conversion of Road Networks from OpenDRIVE to Lanelets},
	booktitle = {Proc. of the IEEE International Conference on Service Operations and Logistics, and Informatics},
	year = {2018},
}

@misc{lgln_basis_dlm_shapefile,
  author       = {{Landesamt für Geoinformation und Landesvermessung Niedersachsen (LGLN)}},
  title        = {{Basis-DLM - Shapefile}},
  year         = {2026},
  url          = {https://ni-lgln-opengeodata.hub.arcgis.com/documents/lgln-opengeodata::basis-dlm-shapefile/about},
  note         = {OpenGeoData portal, accessed May 15, 2026}
}

@misc{lgln_dgm1,
  author       = {{Landesamt für Geoinformation und Landesvermessung Niedersachsen (LGLN)}},
  title        = {{Digitales Geländemodell (DGM1)}},
  year         = {2026},
  url          = {https://ni-lgln-opengeodata.hub.arcgis.com/apps/lgln-opengeodata::digitales-gel%C3%A4ndemodell-dgm1/about},
  note         = {OpenGeoData portal, accessed May 15, 2026}
}

@misc{antlr4_github,
  author       = {{ANTLR Project}},
  title        = {{ANTLR4}},
  year         = {2026},
  howpublished = {\url{https://github.com/antlr/antlr4}},
  note         = {GitHub repository, accessed May 15, 2026}
}

\end{document}